\pdfoutput=1

\documentclass[11pt]{article}

\usepackage[]{acl}

\usepackage{times}
\usepackage{latexsym}
\usepackage{footmisc}

\usepackage[T1]{fontenc}
\usepackage[pdftex]{graphicx}
\usepackage{comment}
\usepackage{stfloats}
\usepackage{tabularx}

\usepackage[utf8]{inputenc}

\usepackage{microtype}

\usepackage{CJKutf8}
\usepackage{color}

\title{Generating Repetitions with Appropriate Repeated Words}

\author{Toshiki Kawamoto, Hidetaka Kamigaito, Kotaro Funakoshi and Manabu Okumura\\
         Tokyo Institute of technology \\ 
         \texttt{\{kawamoto,kamigaito,funakoshi,oku\}@lr.pi.titech.ac.jp}} 

\begin{document}
\maketitle
\begin{abstract}
A repetition is a response that repeats words in the previous speaker's utterance in a dialogue. 
Repetitions are essential in communication to build trust with others, as investigated in linguistic studies.
In this work, we focus on repetition generation. 
To the best of our knowledge, this is the first neural approach to address repetition generation.
We propose Weighted Label Smoothing, a smoothing method for explicitly learning which words to repeat during fine-tuning, and a repetition scoring method that can output more appropriate repetitions during decoding. We conducted automatic and human evaluations involving applying these methods to the pre-trained language model T5 for generating repetitions. The experimental results indicate that our methods outperformed baselines in both evaluations.
\end{abstract}

\section{Introduction}
Dialogues can build a trusting relationship with others, thus are essential in our daily lives  \cite{schein1993dialogue,searle1985speech}.
There are several types of responses in dialogues, and the one we focus on is repetitions \cite{tannen1987repetition}.
A repetition is a response that uses the previous speaker's words or phrases.
Figure \ref{fig:repeat} shows an example. The phrases "a bear" and "came out" are repeated.
Repetitions frequently appear in a conversation with diverse roles, e.g., 
to indicate attentive listening, confirm the previous utterance, and show agreement or sympathy \cite{machi2019managing,shimojima2002informational}.
Many linguistic studies investigating repetitions have concluded that they are important for building and strengthening relationships between speakers \cite{tannen1989talking,johnstone2002discourse,norrick1987functions,brown1999repetition}.
From the above linguistic point of view, we can say that repetitions are indispensable in dialogues.

Repetitions are similar to paraphrases and reflections, which are component skills of counseling \cite{theron2008manual}, in terms of using the previous speaker's utterance. Paraphrases and reflections have been generated using a template-based method \cite{han-etal-2013-counseling}. 

While many studies have tackled general response generation with neural network-based frameworks \cite{adiwardana2020humanlike,zhang-etal-2020-dialogpt}, less attention has been paid to repetitions. This might be because they are buried in a huge amount of response data. 
Therefore, we focus on automatically generating repetitions. To the best of our knowledge, this is the first study on generating repetitions with a neural approach. We used the pre-trained language model T5 \cite{raffel2019exploring} for generating repetitions because it has performed well in language generation in past few years (e.g., \citealp{radfordlanguage, raffel2019exploring, lewis-etal-2020-bart}). 

\begin{figure}[t]
    \centering
    \small
    \begin{tabular}{|l|}
        \hline
        Speaker: When I was driving,\colorbox{yellow}{a bear}suddenly\colorbox{yellow}{came out}. \\
        Listener: Oh.\colorbox{green}{A bear came out}!?\\
        \hline
    \end{tabular}
    \caption{Example repetition. Listener's response uses words from previous speaker's utterance. Yellow words indicate those that are repeated and green words indicate those in the repetition.}
    \label{fig:repeat}
\end{figure}

In generating repetitions, it is important to take into account which words should be repeated from the previous utterance.
The repeated words might represent objective facts, names of people and places, and the speaker's experiences and emotions, though they are different depending on the language \cite{machi2008repetition}. 
When we use a pre-trained language model, however, the model cannot explicitly learn the repeat likelihood among words during fine-tuning because it is difficult to directly teach which words are likely to be repeated at this step.

To solve this problem, we propose Weighted Label Smoothing (WLS), which is an improvement upon Label Smoothing (LS) \cite{szegedy2016rethinking}.
The method enables a language model-based response generator to learn the words it should use for each input utterance during fine-tuning.
We also propose the repetition scoring method (RSM) to expand the scoring method proposed in \citet{wu2016google} for selecting repetitions that contain appropriate repeated words during decoding.

We evaluated the proposed methods on a dataset we created in Japanese for automatic and human evaluations.
Our methods outperformed baselines, i.e., fine-tuned pre-trained language models without our methods, in both evaluations.
This indicates that our methods can generate repetitions that contain appropriate words to repeat.

Our contributions are as follows:
\begin{enumerate}
   \item To the best of our knowledge, this is the first study to use a neural model for generating repetitions.
   \item We will release our code and the dataset of repetitions we created.\footnote{\url{https://github.com/titech-nlp/repetition-generation}}
   \item We propose WLS, that takes into account words that should be repeated during fine-tuning, for generating repetitions.
   \item We propose RSM to select repetitions containing appropriate repeated words during decoding.
\end{enumerate}

\section{Proposed Methods}
Repetitions do not necessarily mean we repeat any word. 
For the utterance "Today's dinner was pizza.", the repetition "Oh, you ate pizza." is more appropriate than "Oh, you ate today."
However, a fine-tuned pre-trained language model alone may not be enough to generate repetitions with appropriate repeated words.
Therefore, to generate a response that repeats more appropriate words, we introduce \textit{repeat scores} (\S \ref{sec:score}) to calculate how likely a word is repeated and incorporate the scores into WLS (\S \ref{sec:weighted}) for fine-tuning and RSM (\S \ref{sec:coverage}) for beam search in decoding.

\subsection{Repeat Score}\label{sec:score}
We should give high scores to words that tend to be used in repetitions and low scores to words that should not be. 
Since only content words (nouns, verbs, adjectives, or adverbs) are repeated in Japanese, we define a \textit{repeat score} only for them. 
Since subwords are used as a unit in a pre-trained language model, all the subwords in the same content word receive the same \textit{repeat score}.

We use BERT \cite{bert} to construct a model for scoring the \textit{repeat scores} in the range of [0, 1]. 
We pass the final hidden state of BERT through SpanExtractor \cite{lee-etal-2017-end} for each word and then convert the vector to a scalar value through a multi-layer perceptron, which has a sigmoid function as the last layer.
In the training data, the label is set to 1 if the target content word was repeated, and 0 if it was not.
The output is then normalized by applying min-max scaling.

\subsection{Weighted Label Smoothing (WLS)}\label{sec:weighted}
In this section, we explain how to learn words to repeat when fine-tuning a pre-trained language model for repetition generation.
Neural response generation models try to optimize cross-entropy loss.
Let $X$ be a previous utterance and $Y$ be a response, where 
$Y$ is divided into subwords as $Y=y_{1}, \dots, y_{T}$.
Letting $K$ be the total number of subwords 
and $v_{k}$ be the $k$-th subword, the cross-entropy loss is defined as follows:
\begin{displaymath}
    L(q, p)=-\sum_{k=1}^{K}q(v_{k})\log\{p(v_{k}|y_{<t},X)\},
\end{displaymath}
where $p(v_{k}|y_{<t},X)$ is the probability of $v_{k}$ that the model outputs at time step $t$ given $X$, and $q(v_{k})$ is the probability of $v_{k}$ in a target distribution that the model aims for.
When a one-hot distribution is used, $q(v_{k})$ is as follows with a function $\delta_{v_{k},y_{t}}$, which becomes 1 when $v_{k}=y_{t}$:
\begin{displaymath}
    q(v_{k}) = \delta_{v_{k},y_{t}}.
\end{displaymath}
When LS is used, however, $q(v_{k})$ is as follows with uniform distribution $u(v_{k})=1/K$:
\begin{displaymath}
    q(v_{k}) = (1-\epsilon)\delta_{v_{k},y_{t}} + \epsilon u(v_{k}),
\end{displaymath}
where $\epsilon$ is a hyperparameter.

A one-hot distribution and LS cannot learn a subword to repeat explicitly because there are labels other than the target, i.e., $v_{k}$ when $v_{k} \neq y_{t}$, that have the same $q(v_{k})$.
Therefore, we propose WLS, which takes into account how likely a subword is repeated.
We use \textit{repeat scores}, explained in \S \ref{sec:score}, instead of $u(v_{k})$.
The $q(v_{k})$ of WLS is defined as follows:
\begin{displaymath}
    q(v_{k}) = (1-\epsilon)\delta_{v_{k},y_{t}} + \epsilon \frac{r(v_{k})^\gamma}{K},
\end{displaymath}
where $r(v_{k})$ is the \textit{repeat score} for $v_{k}$, and $\gamma$ is a hyperparameter.
We use the $q(v_{k})$ of WLS as the distribution in the cross-entropy loss function.
Subwords in the previous speaker's utterance are weighted in accordance with their $r(v_{k})$. 
Note that if we set $\gamma = 0$, WLS is the same as LS.

\begin{table*}[t]
    \centering
    \small
    \begin{tabular}{l|l}
    \hline
        Dialogue & \begin{tabular}{l}
            \begin{CJK}{UTF8}{ipxm}運動やってました. \end{CJK}\\
            (I played sports.)\\
            \begin{CJK}{UTF8}{ipxm}へーそしたら中学校高校はクラブは何か？\end{CJK}\\
            (Oh, did you participate in any clubs in junior high or high school?)\\
            \begin{CJK}{UTF8}{ipxm}中学高校大学まで陸上部でした. \end{CJK}\\
            (I was a member of a track and field club in junior high, high school, and college.)\\
        \end{tabular}\\ \hline \hline
        Repetition1 & \begin{tabular}{l}
            \begin{CJK}{UTF8}{ipxm}陸上部ですか. \end{CJK}\\
            (Track and field club?)\\
        \end{tabular}\\
        \hline
        Repetition2 & \begin{tabular}{l}
            \begin{CJK}{UTF8}{ipxm}陸上部だったんですね. \end{CJK}\\
            (You were in the track and field club.)\\
        \end{tabular}\\
        \hline
        Repetition3 & \begin{tabular}{l}
            \begin{CJK}{UTF8}{ipxm}中学高校大学まで陸上とは, 長く続けられたんですね！\end{CJK}\\
            (You were in the track and field club for a long time, from junior high through high school and college.)\\
        \end{tabular}\\
    \hline
    \end{tabular}
    \caption{Examples from repetition dataset. There are at most three repetitions for one dialogue.}
    \label{tab:data_example}
\end{table*}

\subsection{Repetition Scoring Method (RSM)}\label{sec:coverage}
Pre-trained language models usually use beam search in decoding.
We propose a scoring method, RSM, to select more appropriate repetitions in the beam search.
RSM is an extension of a scoring method for machine translation in \citet{wu2016google}.
The original scoring method uses a length normalization procedure and coverage penalty \cite{tu-etal-2016-modeling}.
Length normalization treats sentences of different lengths equally.
The coverage penalty gives a high score to a sentence that is most likely to cover all the words in the source sentence.
Since the original scoring method cannot select a repetition with appropriate repeated words, 
we modify the method by adding \textit{repeat scores}, which indicate words to repeat.
Letting $Y$ be a candidate response during beam search and $X$ be the previous utterance, the generation probability is $P(Y|X)$.
The scoring function $s(Y,X)$ of RSM is as follows:

\footnotesize
\begin{eqnarray*}
    s(Y,X)\!&=&\!log\{P(Y|X)\}/lp(Y)\mathalpha{+}cp(Y,\!X)\mathalpha{+}rs(Y,\!X),\\
    lp(Y)\!&=&\!\frac{(5+|Y|)^\alpha}{(5+1)^\alpha},\\
    cp(Y,X)\!&=&\!\beta*\sum^{|X|}_{i=1}log(\sum^{|Y|}_{j=1}p_{i,j}),\\
    rs(Y,X)\!&=&\!log\sum^{|Y|}_{j=1}r(v_{j}),
\end{eqnarray*}
\normalsize
where $\alpha$ and $\beta$ are hyperparameters for length normalization and coverage penalty, respectively. 
We carry out two modifications to the original scoring method to yield RSM.
First, we use the attention value of $p_{i,j}$ without suppression.
In contrast to machine translation, in which an input and output have a one-to-one relationship, lengths of an input and output are not the same in repetition generation, and so it is not suitable to suppress the attention value under $1.0$.
Second, we add the term $rs(Y,X)$, which represents the sum of \textit{repeat scores} for subwords in the response.

\section{Dataset}\label{sec:data}
We manually created pairs of a speaker’s utterance and its repetition as our dataset using a crowdsourcing service.\footnote{\url{https://www.lancers.jp/}\label{fo:lancers}}
Since repetitions often occur when a listener replies to a speaker, we used utterances in a corpus of listening dialogues \cite{corpus} between an elderly person and caregiver or clinical psychologist as the speaker's utterances in our dataset.\footnote{We attempted to extract repetitions from the corpus using a rule-based approach and found it is difficult to obtain a sufficient amount of such repetitions.}
In this corpus, the elderly person tends to be a speaker and the others are listeners.
We extracted the elderly person's utterances containing content words for creating a repetition.
The number of extracted utterances was 5,548.
We asked three crowdsourcing workers to create repetitions for each utterance.
Specifically, a worker was shown two utterances before each target utterance and asked to create a repetition, that supports the creation of context-aware repetitions.
When the workers found it difficult to create a repetition for an utterance, they could discard it.
The total number of workers was 333.

Examples from the dataset are given in Table \ref{tab:data_example}.
The size and statistics of our repetition dataset are shown in Tables \ref{tab:dataset} and \ref{tab:stats_dataset}.
The word overlap rate is the percentage of words in an utterance that are repeated in a repetition. 
The content-word overlap rate is the percentage of content words of an utterance that are repeated. 
Comparing the average numbers of tokens, repetitions are much shorter than utterances. 
This may indicate that repetitions cannot be produced simply by copying the utterances, and we need to select information that is worth repeating from the utterances.

To understand what types of words overlap, Table \ref{tab:pos} shows the percentage of all words' parts-of-speech and overlapped words' parts-of-speech in utterances. 
Since "postpositional particles" and "auxiliary verbs" tend to accompany content words in a Japanese unit called `bunsetsu', 
it might be natural that they also appear in repetitions in high percentages.

While we can have at most three repetitions for an utterance in our dataset, we used only one randomly selected repetition for an utterance in the training data. We used all repetitions for an utterance for the evaluation on the validation and test data to consider the diversity of responses.

\begin{table}[t]
    \centering
    \small
    \begin{tabular}{l|r|r|r}
        \hline
        &Train.&Valid.&Test\\
        \hline
        Utterance (Dialogue) & 4106 & 489 & 490\\
        Repetition & 10677 & 1305 & 1312\\
        \hline
    \end{tabular}
    \caption{Size of repetition dataset.}
    \label{tab:dataset}
\end{table}

\begin{table}[t]
    \centering
    \small
    \begin{tabular}{lr}
        \hline 
         Total Dialogues & 5085\\
         Total Repetitions & 13294\\
         Average \# of Repetitions per Utterance & 2.61\\
         Average \# of Tokens per Utterance & 26.25\\
         Average \# of Tokens per Repetition & 11.74\\
         Word Overlap Rate & 36.48\%\\
         Content-word Overlap Rate & 38.14\%\\
        \hline
    \end{tabular}
    \caption{Statistics of repetition dataset.}
    \label{tab:stats_dataset}
\end{table}

\begin{table}[t]
    \centering
    \small
    \begin{tabular}{l|r|r}
        \hline 
        PoS & All(\%) & Overlap(\%)\\
        \hline
        Postpositional Particle & 27.64 & 39.02\\
        Noun & 23.85 & 32.70\\
        Auxiliary Verb & 9.34 & 13.09\\
        Verb & 13.25 & 10.03\\
        Adjective & 1.86 & 2.52\\
        Adverb & 4.57 & 1.61 \\
        Filler & 0.37 & 0.01\\
        \hline
    \end{tabular}
    \caption{The ratios of words and overlapped words of different parts-of-speech (PoS) in utterances.}
    \label{tab:pos}
\end{table}

\section{Experiments}
\subsection{General Setup}
\textit{Repeat scores} were calculated from the training data. 
SentencePiece \cite{kudo-richardson-2018-sentencepiece} was used to segment the dataset into subwords.
With WLS, the hyperparameter $\epsilon$ was set to $0.1$ following a previous study \cite{szegedy2016rethinking}, and $\gamma$ was tuned to $4$ with the validation data, as explained in Appendix \ref{sec:hyper}. 
With RSM, we used $\alpha=0.2$ and $\beta=0.2$, following a previous study \cite{wu2016google}, and a beam size of 5.
We used MeCab\footnote{\url{https://taku910.github.io/mecab/}\label{mecab}} as a tokenizer to identify content words. 

\subsection{Compared Methods}
The baseline methods were as follows:

\begin{table}[t]
    \centering
    \scriptsize
    \begin{tabular}{ll}
        \hline
        Utterance&\hspace{-1mm}Rule-Based\\
        \hline
        \hspace{-2mm}\begin{tabular}{l}
            \begin{CJK}{UTF8}{ipxm}それとやっぱり深さ, 魚のどの辺におるか\end{CJK}\\
            \begin{CJK}{UTF8}{ipxm}とかが難しいんですわ. \end{CJK}\\
            (It's hard to know where fish are and what\\
            depths they are at.)\\
		\end{tabular}&\hspace{-3mm}\begin{tabular}{l}
            \begin{CJK}{UTF8}{ipxm}難しいですか. \end{CJK}\\
            (Hard, is it?)\\
		\end{tabular}\\
        \hline
        \hspace{-2mm}\begin{tabular}{l}
            \begin{CJK}{UTF8}{ipxm}先生の話をしっかり聞く言う事が大事. \end{CJK}\\
            (It is important to listen carefully to what\\
            a teacher says.)\\
		\end{tabular}&\hspace{-3mm}\begin{tabular}{l}
            \begin{CJK}{UTF8}{ipxm}先生ですか. \end{CJK}\\
            (The teacher, is it?)\\
		\end{tabular}\\
		\hline
		\hspace{-2mm}\begin{tabular}{l}
            \begin{CJK}{UTF8}{ipxm}色々ありましたからね, 国際的なニュース\end{CJK}\\
            \begin{CJK}{UTF8}{ipxm}もね. \end{CJK}\\
            (There's been a lot going on, and international\\
            news, too.)\\
		\end{tabular}&\hspace{-3mm}\begin{tabular}{l}
            \begin{CJK}{UTF8}{ipxm}ニュースですか. \end{CJK}\\
            (News, is it?)\\
		\end{tabular}\\
        \hline
    \end{tabular}
    \caption{Examples of utterance and rule-based response.}
    \label{tab:rule}
\end{table}

\noindent\textbf{Rule-Based} is a rule-based method, with which a response is created with a content word in the speaker's utterance + \textit{"desuka"} ("is it?").
The content word is randomly selected from the utterance.
Examples of rule-based responses are given in Table \ref{tab:rule}.
Responses made with Rule-Based always contain a repeated word and have few grammatical errors.
However, \textit{"desuka"} cannot cover all situations.
\textit{"desuka"} was chosen because 52\% of repetitions in our dataset ends with \textit{"desuka"}, and 6.1\% of repetitions are a single word + \textit{"desuka"}.

\noindent\textbf{BertSumAbs} \cite{liu-lapata-2019-text} is a model trained with BERT\footnote{\url{https://huggingface.co/cl-tohoku/bert-base-japanese}} as the encoder and randomly initialized Transformer as the decoder.

\noindent\textbf{T5}\footnote{\url{https://huggingface.co/sonoisa/t5-base-japanese}} \cite{raffel2019exploring} is a model that was fine-tuned with the repetition dataset.\footnote{While another possible model for comparison is GPT-2 \cite{radfordlanguage}, we did not use it since it was known that T5 is superior to GPT-2 in generation performance \cite{kale-rastogi-2020-text,zhao-etal-2020-knowledge-grounded}.}

\noindent\textbf{LS} is T5 fine-tuned with LS.

\noindent\textbf{Copy} is T5 fine-tuned with the copy mechanism \cite{see-etal-2017-get}.
Since the copy mechanism can be considered similar to the repetition model in that it is used to generate the same words as in an input sentence, we used it for comparison.

Note that the T5 and BERT were versions pre-trained in Japanese.
Our methods are as follows:

\noindent\textbf{WLS} is T5 fine-tuned with WLS, as mentioned in \S \ref{sec:weighted}. 

\noindent\textbf{RSM} is T5 using RSM during beam search, as mentioned in \S \ref{sec:coverage}.

\noindent\textbf{WLS + RSM} is T5 fine-tuned with WLS and using RSM during beam search.

\subsection{Automatic Evaluation}
The evaluation metrics were ROUGE (RG-1, RG-2, RG-L) \cite{lin-2004-rouge} and the percentage of outputs containing correct repeated words.
The correct repeated words are content words repeated in the gold response.
The experimental results are listed in Table \ref{tab:result}.
WLS + RSM obtained the highest scores for all metrics, confirming the effectiveness of both WLS and RSM.

\begin{table}[t]
    \centering
    \small
    \begin{tabular}{l|c|c|c|c}
    \hline
     & RG-1 & RG-2 & RG-L & \% \\
    \hline 
    Rule-Based & 35.26 & 14.03 & 35.11 & 58.24 \\
    BertSumAbs & 30.73 & 10.97 & 29.94 & 52.51 \\
    T5 & 45.34 & 22.34 & 44.59 & 81.67 \\
    LS & 45.89 & 23.08 & 45.12 & 81.83 \\
    Copy & 45.83 & 23.32 & 45.07 & 81.67 \\
    \hline
    WLS & \hspace{1.1mm}\,47.88$^\dag$ & 24.56 & \hspace{1.1mm}\,47.14$^\dag$ & \hspace{1.1mm}\,85.77$^\dag$ \\
    RSM & 46.96 & 24.66 & 46.13 & \hspace{1.1mm}\,84.38$^\dag$ \\
    WLS + RSM & \hspace{1.1mm}\,\textbf{49.16}$^\dag$ & \hspace{1.1mm}\,\textbf{26.58}$^\dag$ & \hspace{1.1mm}\,\textbf{48.28}$^\dag$ & \hspace{1.1mm}\,\textbf{89.56}$^\dag$ \\
    \hline
    \end{tabular}
    \caption{Results of automatic evaluation. \% is percentage of outputs containing correct repeated words. Results with \dag \hspace{0.5mm} are significantly different from LS, best baseline, using Wilcoxon rank sum test (p < 0.05).}
    \label{tab:result}
\end{table}

We conducted an ablation study to analyze the results of RSM.
The results are listed in Table \ref{tab:cov_ablation}.
Since w/o $rs$ received the lowest scores, $rs$ was considered the most effective.

\begin{table}[t]
    \centering
    \small
    \begin{tabular}{l|c|c|c|c}
        \hline
        & RG-1 & RG-2 & RG-L & \% \\
        \hline 
        w/o $lp$  & 46.81 & 24.52 & 45.98 & 83.91\\
        w/o $cp$  & 46.93 & \textbf{24.66} & 46.11 & 84.30\\
        w/o $rs$  & 44.97 & 22.25 & 44.13 & 81.28\\
        \hline 
        RSM & \textbf{46.96} & \textbf{24.66} & \textbf{46.13} & \textbf{84.38}\\
        \hline
    \end{tabular}
    \caption{Ablation study for RSM. $lp$, $cp$, and $rs$ were explained in \S \ref{sec:coverage}.}
    \label{tab:cov_ablation}
\end{table}

Examples of an input and generated responses from the baseline and our model are shown in Table \ref{tab:res_example}.
The proposed model (WLS + RSM) successfully generated a response that was close to the correct response, focusing on "having friends who play Go".

\begin{table}[t]
    \centering
    \scriptsize
    \begin{tabular}{l|l}
    \hline
        Utterance & \begin{tabular}{l}
            \begin{CJK}{UTF8}{ipxm}昨日は同じ僕らの仲間で囲碁する人いたからね. \end{CJK}\\
            (Yesterday there were our friends who play Go.)\\
        \end{tabular}\\ \hline\hline
        Gold & \begin{tabular}{l}
            \begin{CJK}{UTF8}{ipxm}仲間で囲碁する人いたんですね. \end{CJK}\\
            (There were friends who play Go.)\\
        \end{tabular}\\ \hline
        Rule-Based & \begin{tabular}{l}
            \begin{CJK}{UTF8}{ipxm}囲碁ですか. \end{CJK}\\
            (Go, is it?)\\
        \end{tabular}\\ \hline
        
        T5 & \begin{tabular}{l}
            \begin{CJK}{UTF8}{ipxm}囲碁をしてくれたんですね. \end{CJK}\\
            (You played Go.)\\
        \end{tabular}\\ \hline
        Ours & \begin{tabular}{l}
            \begin{CJK}{UTF8}{ipxm}仲間内で囲碁する人いたんですか. \end{CJK}\\
            (There were friends who play Go.)\\
        \end{tabular}\\
    \hline
    \end{tabular}
    \caption{Examples of generated responses from different models.}
    \label{tab:res_example}
\end{table}

\subsection{Human Evaluation}
We also conducted a human evaluation by comparing three types of response generation methods: Rule-Based, T5, and our model (WLS + RSM).
The evaluation measures were grammaticality (Gram), relevance (Rel), coherence (Cohe), and whether repeated words are included (Rep).
Two hundred pairs were randomly selected from the test data. The responses were shown to five workers and evaluated on a three-point Likert scale.
The response was evaluated with the previous speaker's utterance and one turn before the speaker's utterance as context, meaning the context helps in determining whether the response is an appropriate repetition.
The total number of evaluators was 110.

Average scores from the evaluation are listed in Table \ref{tab:human_eva}.
WLS + RSM outperformed the other methods for all measures, confirming its effectiveness.

\begin{table}[t]
    \centering
    \small
    \begin{tabular}{l|c|c|c|c}
        \hline 
         & Gram & Rel & Cohe & Rep\\
        \hline 
         Rule-Based & 2.63 & 2.49 & 2.37 & 2.64\\
        T5 & 2.82 & 2.77 & 2.62 & 2.79\\
        WLS + RSM & \hspace{1.1mm}\,\textbf{2.85}$^\dag$ & \hspace{1.1mm}\,\textbf{2.80}\hspace{1.1mm}\, & \hspace{1.1mm}\,\textbf{2.64}\hspace{1.1mm}\, & \hspace{1.1mm}\,\textbf{2.88}$^\dag$ \\
        \hline
    \end{tabular}
    \caption{Results of human evaluation. Results with \dag \hspace{1mm} are significantly different from T5, the best baseline, using Wilcoxon rank sum test (p < 0.05).}
    \label{tab:human_eva}
\end{table}

\section{Conclusion}
We focused on repetition generation.
Although repetitions play an important role in dialogues, there has been no neural approach for this task to the best of our knowledge.
We proposed WLS, which is an improvement upon LS, during fine-tuning and RSM, which is an extended scoring method, during decoding for repetition generation.
Through automatic and human evaluations, we confirmed that our model can generate repetitions that contain more appropriate words to repeat than baseline models.
For future work, we will take into account synonyms and multiple gold repetition instances to calculate \textit{repeat scores} for improving the diversity of responses. We are also planning to incorporate our repetition model into a general response generation framework.

\section*{Acknowledgements}
We thank Dr. Koichiro Yoshino of RIKEN for providing the listening dialogue data.

\section*{Ethical Statement}
Neural generative models have the potential to generate unexpected responses such as violent remarks. As we focused on repetition generation, its model repeats a user's utterance, and so there is little chance of causing unintended responses compared with chit-chat dialogue systems. However, this does not mean that unintended responses will never appear, e.g., when a user's utterance is an unintended expression. Thus, the same consideration must be taken as with other dialogue systems.

Our dataset was created to repeat from utterances in a privacy-secured dataset, and so there is no privacy issue. Since the license of the original dataset is CC BY-NC 4.0, we could use it for this study. We define that the license of our dataset is also CC BY-NC 4.0.

\bibliography{custom}
\bibliographystyle{acl}

\clearpage
\appendix

\begin{table*}[ht]
    \centering
    \small
    \begin{tabular}{c|c|c|c|c|c|c|c|c|c}
    \hline
        $\gamma$ & 0.0 & 0.1 & 0.5 & 1.0 & 2.0 & 3.0 & 4.0 & 5.0 & 10.0  \\
         \hline
        \% & 82.06 & 82.06 & 81.28 & 83.37 & 82.21 & 83.52 & 85.07 & 82.90 & 84.53 \\ \hline
    \end{tabular}
    \caption{Percentage of generated responses containing a correct repeated word in the development data when $\gamma$ was changed. $\epsilon=0.1$. $\gamma=0$ indicates LS. The best score was obtained when $\gamma=4.0$.}
    \label{tab:gamma}
\end{table*}

\section{Exploring Hyperparameter $\gamma$} \label{sec:hyper}
We explored the effect of $\gamma$ on the percentage of responses containing a correct repeated word. 
The model we used for experiments was the pre-trained model T5, fine-tuned with the training data in \S \ref{sec:data}.
We generated repetitions on the development data. 
The results are listed in Table \ref{tab:gamma}. The best score was recorded when $\gamma=4.0$. 
Therefore, we used this value.

\section{P-values} \label{sec:pvalue}
We now discuss the p-values in the experimental results. 
To obtain p-values, we conducted the Wilcoxon rank sum test to compare the effectiveness between baseline models and our proposed models.
Table \ref{tab:result_p} shows the p-values for Table \ref{tab:result} from LS. 
Table \ref{tab:human_eva_p} shows those for Table \ref{tab:human_eva} from T5.
\newpage

\begin{table}[t]
    \centering
    \small
    \begin{tabular}{l|c|c|c|c}
    \hline
     & RG-1 & RG-2 & RG-L & \% \\
    \hline 
    WLS & 0.026 & 0.159 & 0.031 & 0.003 \\
    RSM & 0.225 & 0.141 & 0.294 & 0.047 \\
    WLS + RSM & 0.000 & 0.001 & 0.001 & 0.000 \\
    \hline
    \end{tabular}
    \caption{P-values in Wilcoxon rank sum test between LS and our proposed models in Table \ref{tab:result}.}
    \label{tab:result_p}
\end{table}

\begin{table}[t]
    \centering
    \small
    \begin{tabular}{l|c|c|c|c}
        \hline 
         & Gram & Rel & Cohe & Rep \\
        \hline 
        WLS + RSM & 0.000 & 0.055 & 0.170 & 0.000\\
        \hline
    \end{tabular}
    \caption{P-values in Wilcoxon rank sum test between T5 and WLS + RSM in Table \ref{tab:human_eva}.}
    \label{tab:human_eva_p}
\end{table}

\quad

\end{document}